\documentclass[runningheads]{llncs}

\usepackage{graphicx}
\usepackage[square,sort,comma,numbers,sectionbib]{natbib}
\usepackage[small]{caption}
\usepackage{subcaption}
\usepackage{xcolor}
\usepackage{bbm}
\usepackage{amssymb}
\usepackage{amsmath}
\usepackage{booktabs}
\usepackage{algorithm}
\usepackage{algorithmic}
\usepackage{array}
\usepackage{url}

\newcommand{\keypoint}[1]{\textbf{#1}\quad}
\newcommand{\cut}[1]{}

\newcolumntype{H}{>{\setbox0=\hbox\bgroup}c<{\egroup}@{}}
\begin{document}
%
\title{On Context Distribution Shift in Task Representation Learning for Offline Meta RL}
%
%
\author{Chenyang Zhao \and
Zihao Zhou \and
Bin Liu\thanks{Corresponding author. This paper has been accepted by 19th Inter. Conf. on Intelligent Computing (ICIC 2023).}}
\authorrunning{C. Zhao et al.}

\institute{Research Center for Applied Mathematics and Machine Intelligence, \\Zhejiang Lab, Hangzhou 311121, China\\
\email{\{c.zhao, zhouzihao, liubin\}@zhejianglab.com}\\
}
\maketitle              
\begin{abstract}
Offline Meta Reinforcement Learning (OMRL) aims to learn transferable knowledge from offline datasets to enhance the learning process for new target tasks. Context-based Reinforcement Learning (RL) adopts a context encoder to expediently adapt the agent to new tasks by inferring the task representation, and then adjusting the policy based on this inferred representation. In this work, we focus on context-based OMRL, specifically on the challenge of learning task representation for OMRL. We conduct experiments that demonstrate that the context encoder trained on offline datasets might encounter distribution shift between the contexts used for training and testing. To overcome this problem, we present a hard-sampling-based strategy to train a robust task context encoder. Our experimental findings on diverse continuous control tasks reveal that utilizing our approach yields more robust task representations and better testing performance in terms of accumulated returns compared to baseline methods. Our code is available at \url{https://github.com/ZJLAB-AMMI/HS-OMRL}.
\end{abstract}
\section{Introduction}
\label{sec: intorduction}

Reinforcement learning (RL) has emerged as a powerful technique, having demonstrated remarkable success in several domains such as video games \cite{mnih2015human, berner2019dota}, robotics \cite{haarnoja2018soft}, and board games \cite{silver2017mastering}. However, RL still confronts the challenge of acquiring a satisfactory model or policy in a new setting, which requires a large number of online interactions. This challenge is significant, especially in scenarios where the cost or safety associated with interacting with the environment is high, like health care systems \cite{gottesman2019guidelines} and autonomous driving \cite{zhou2020smarts}.

\begin{figure}[t]
    \centering
    \includegraphics[width=0.78\textwidth]{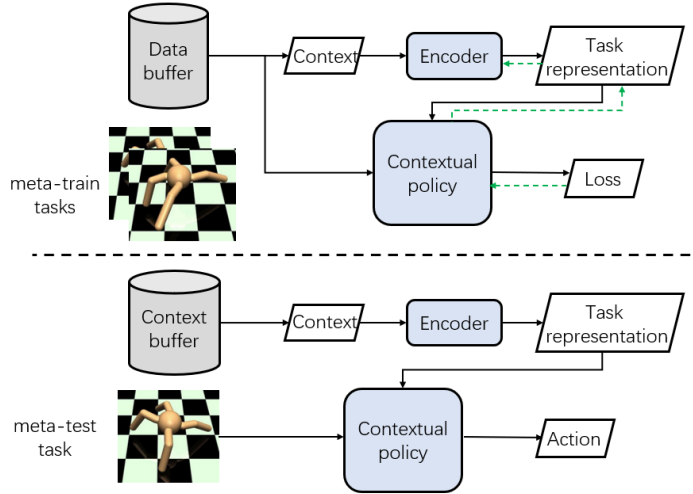}
    \caption{A general framework of context-based meta RL. At the meta-train stage, from the same data buffer, the agent learns to infer about the task and to act optimally in meta-train environments through backpropagation. At the meta-test stage, the agent predicts the task representation with few-shot of context information and adapts the contextual policy through task representation. Solid lines represent forward pass and dash lines represent backward pass.}
    \label{fig:my_label}
\end{figure}

Meta RL has been demonstrated as a promising approach to tackle this issue by learning transferable knowledge about the learning process itself and extracting a meta-policy, which enables rapid adaptation with few samples in unseen target environments \cite{finn2017model, rakelly2019efficient}. Meta RL operates on an assumption that all environments share a similar structure, and it learns about the shared structure by interacting with a distribution of training tasks. Recent research suggests that a meta RL agent can learn to infer about the task from few sample interactions and adjust its policy accordingly \cite{rakelly2019efficient, humplik2019meta, duan2016rl}. Context-based meta RL approaches involve learning a universal policy conditioned on a latent task representation \cite{rakelly2019efficient}. During the meta-test stage, the agent adapts the acting policy based on the predicted task representation through a few online interactions. Fig.~\ref{fig:my_label} illustrates the general framework of context-based meta RL during both meta-train and meta-test stages.

\begin{figure*}[t]
\centering
\begin{subfigure}[b]{0.3\textwidth}
\includegraphics[width=\textwidth]{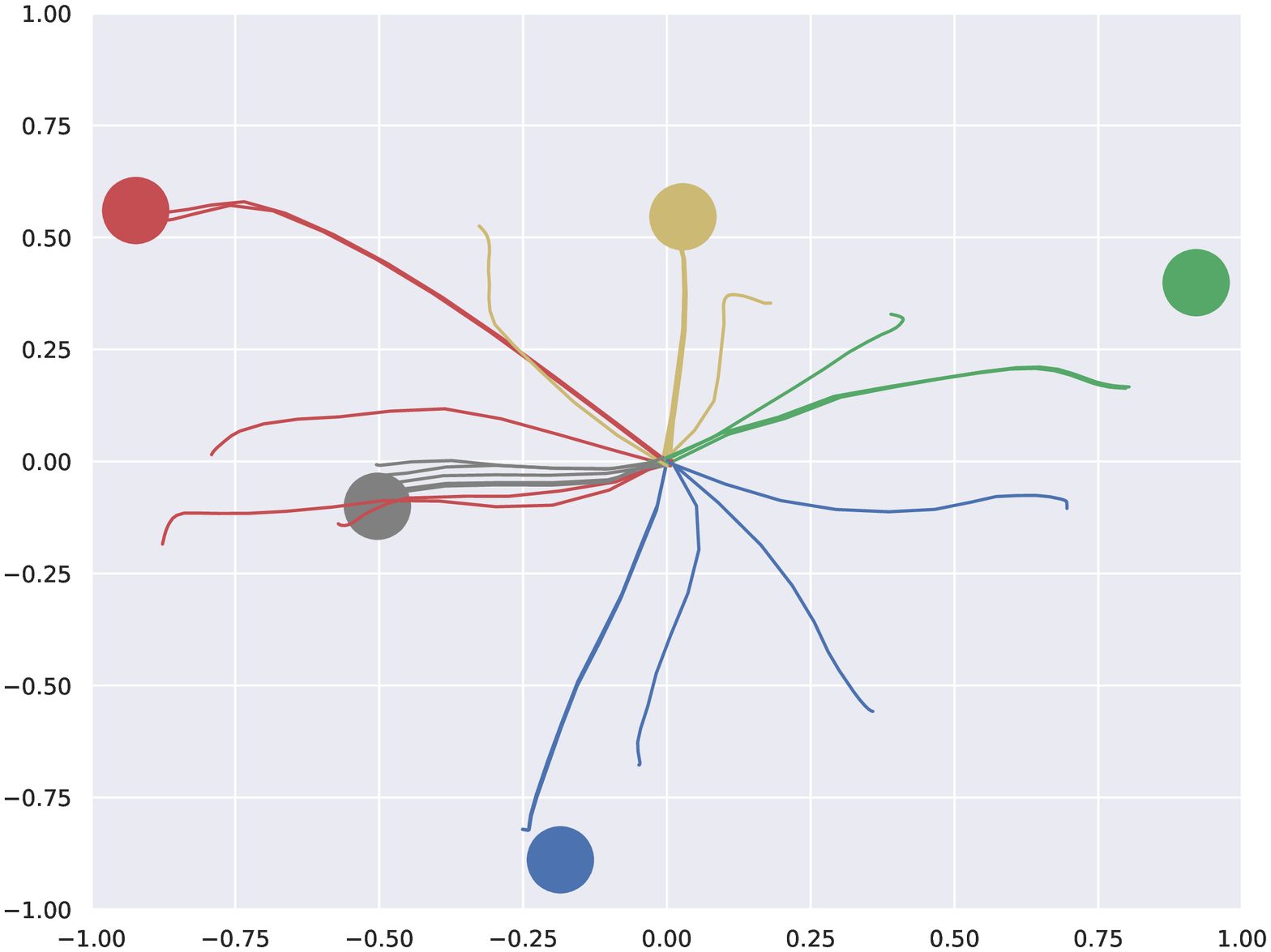}
\end{subfigure}
\begin{subfigure}[b]{0.3\textwidth}
\includegraphics[width=\textwidth]{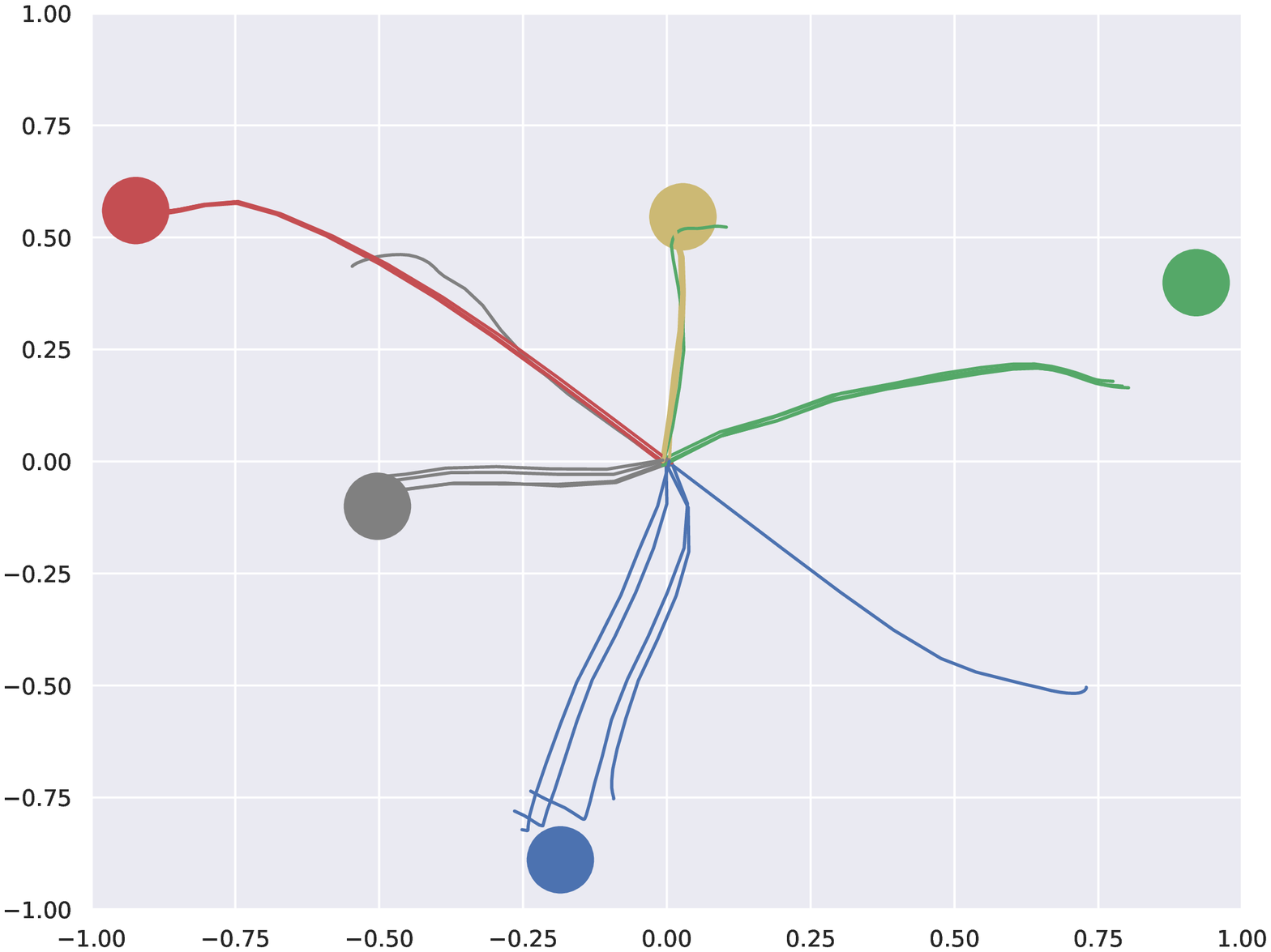}
\end{subfigure}
\begin{subfigure}[b]{0.3\textwidth}
\includegraphics[width=\textwidth]{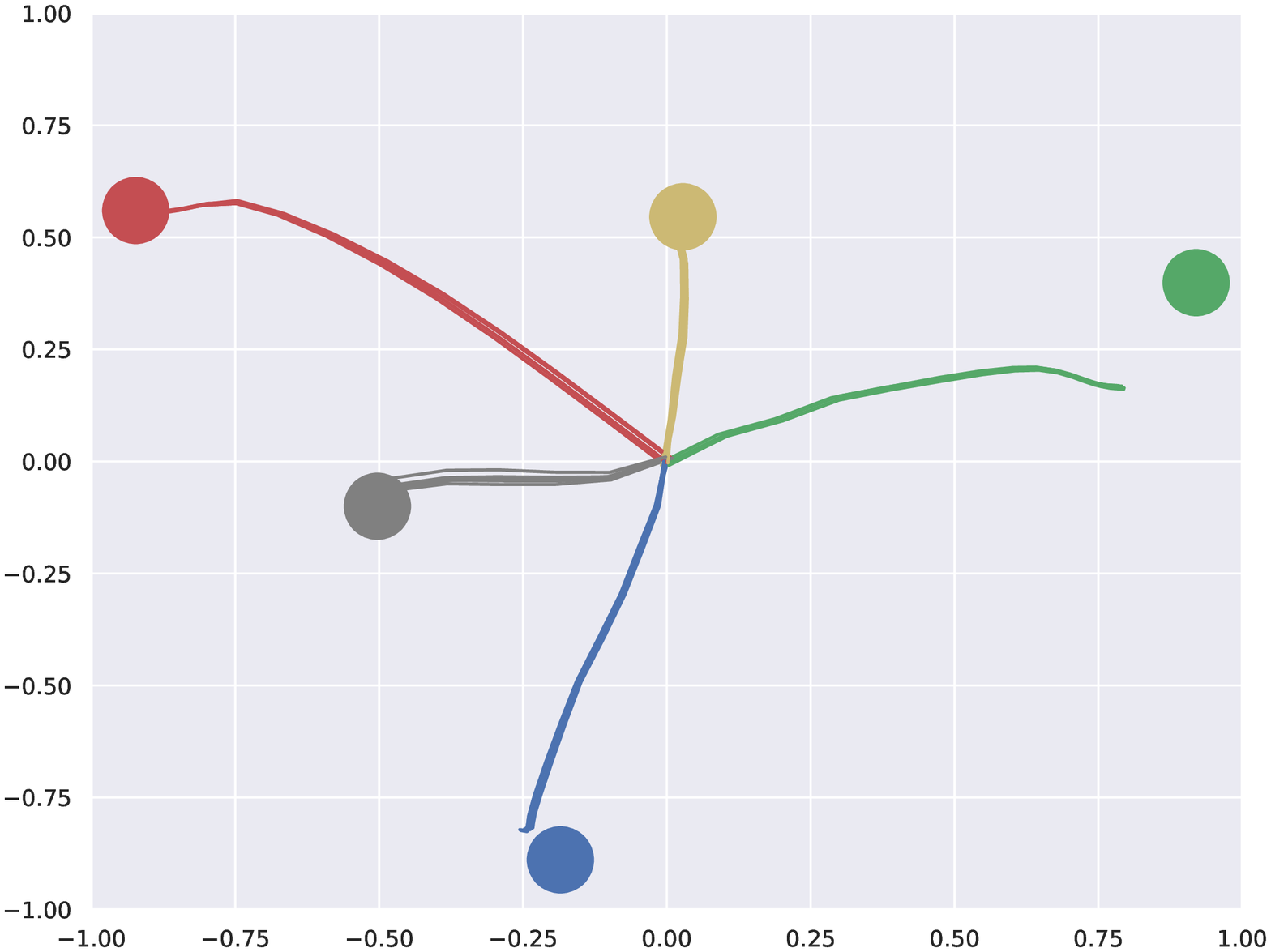}
\end{subfigure}
\\
\begin{subfigure}[b]{0.3\textwidth}
\includegraphics[width=\textwidth]{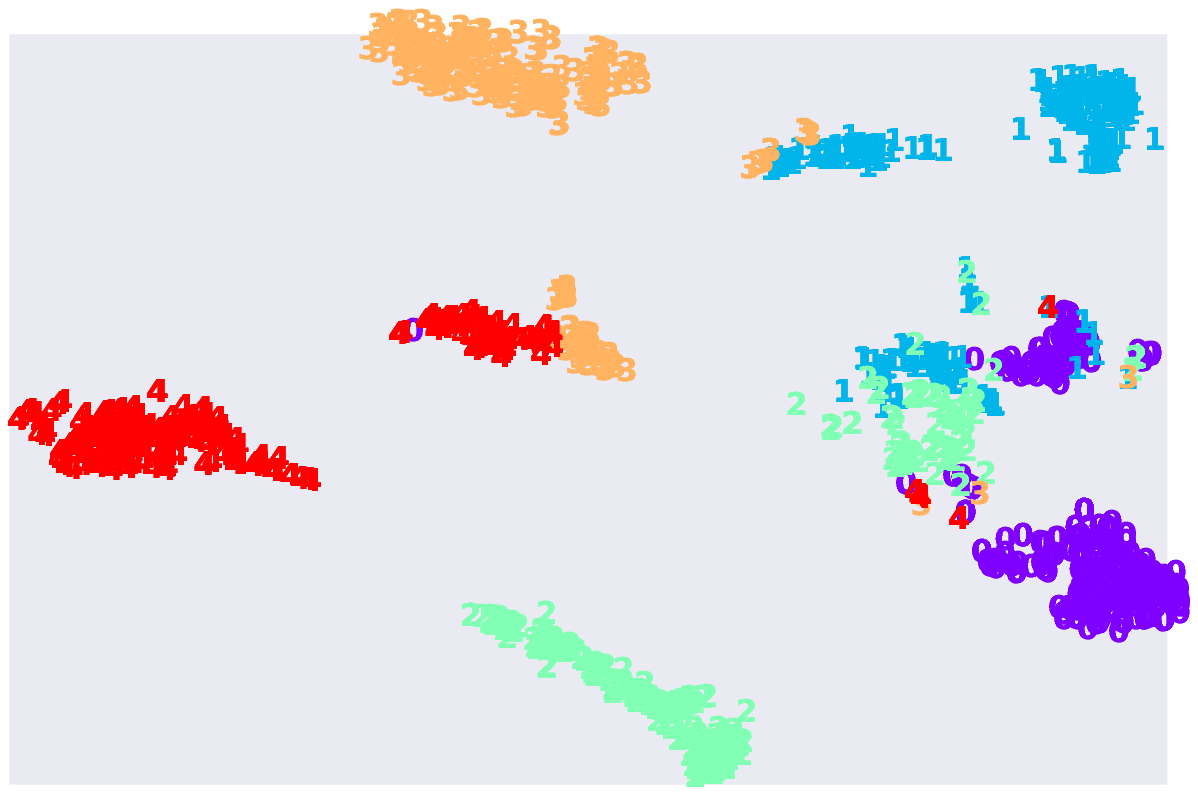}
\caption{Low quality contexts }
\end{subfigure}
\begin{subfigure}[b]{0.3\textwidth}
\includegraphics[width=\textwidth]{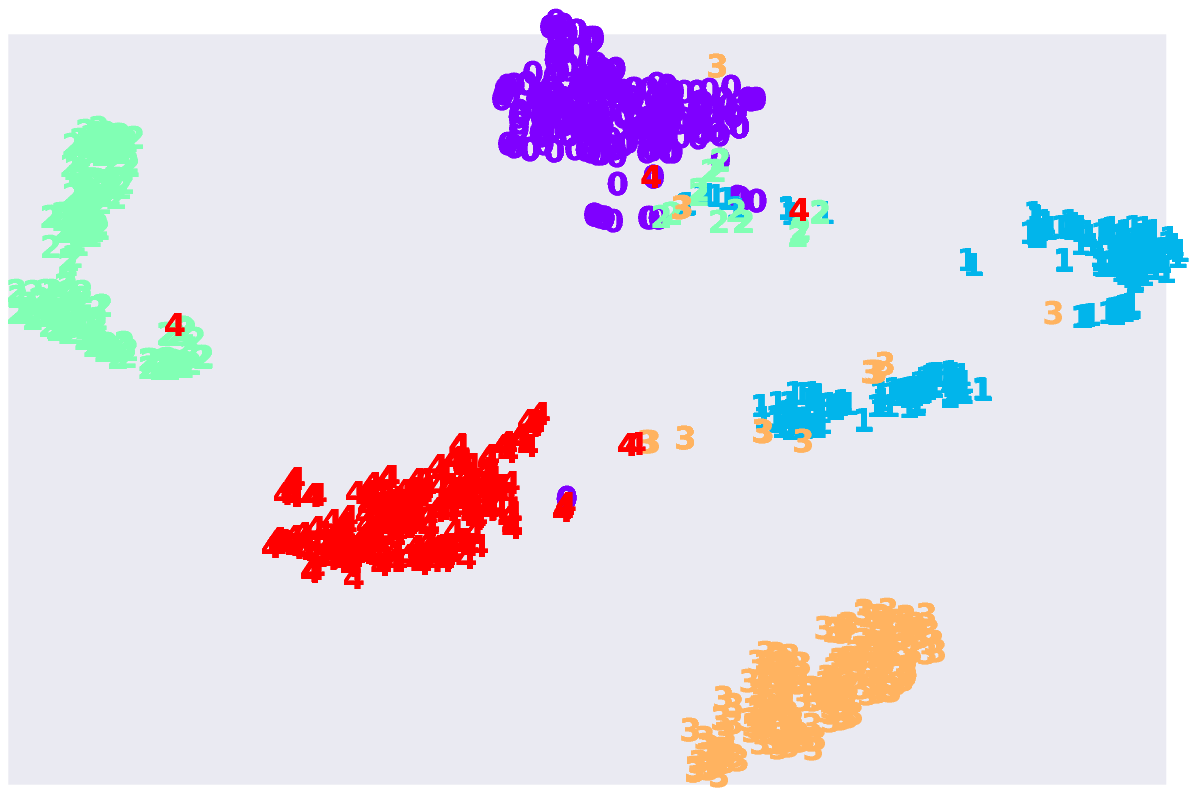}
\caption{Medium quality contexts }
\end{subfigure}
\begin{subfigure}[b]{0.3\textwidth}
\includegraphics[width=\textwidth]{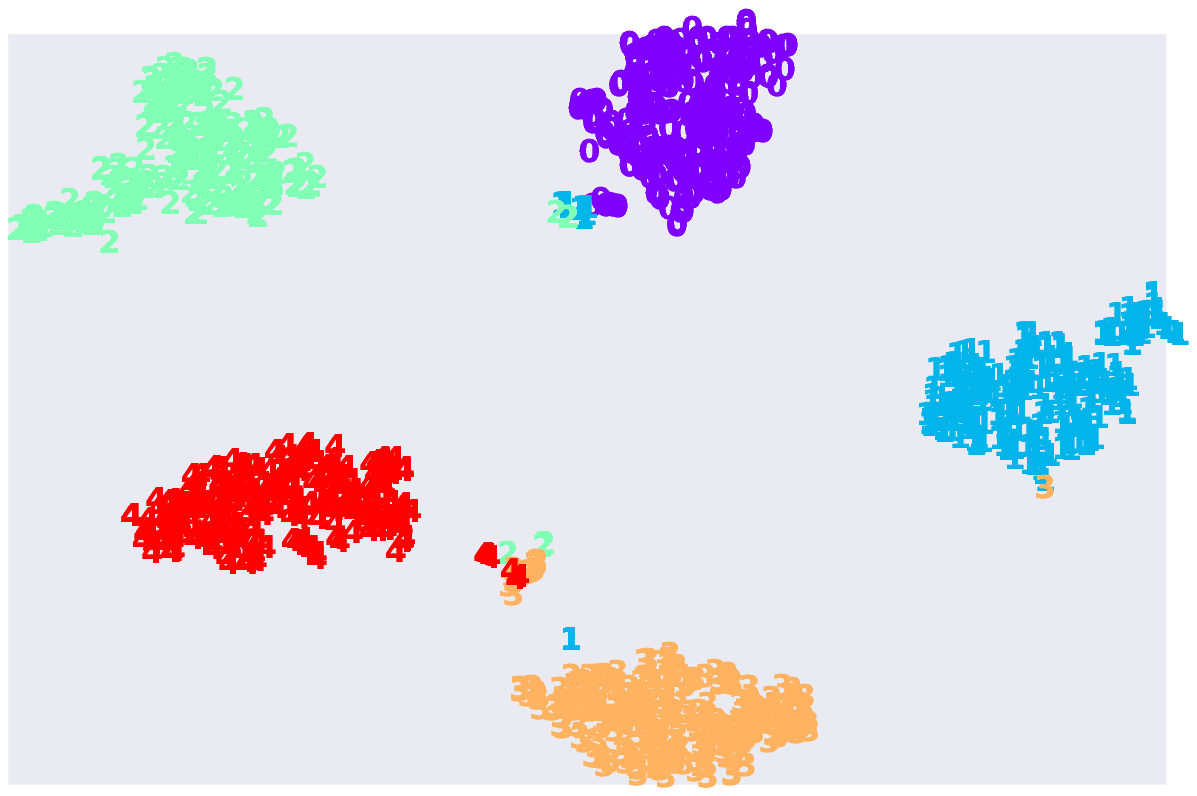}
\caption{High quality contexts }
\end{subfigure}
\caption{Top row: Demonstrations of outcome testing trajectories with contexts of different qualities. Bottom row: t-SNE visualization of the embedded contexts of different qualities. Five random testing goals (colorred dots) are randomly sampled from the task distribution and five contexts are randomly sampled for each goal for visualizing the trajectories. These comparisons show that the performance of the learned context encoder and contextual policy largely depends on the quality of test contexts.}
\label{fig: quality}

\end{figure*}

Although meta RL methods improve sample efficiency during the meta-test stage, a large batch of online experiences is usually required during the meta-train stage, which is collected by intensive interaction with the training environment. Recent research \cite{mitchell2021offline,li2020focal,dorfman2021offline} has combined offline RL ideas \cite{levine2020offline, prudencio2022survey} with meta RL to address data collection problems during meta-training. Li et al. \cite{li2021provably} and Zhou et al. \cite{yuan2022robust} have trained context encoders through supervised contrastive learning, treating same-task samples as positive pairs and all other samples as negative ones. They learned data representations by clustering positive samples and pushing away negative ones in the embedding space. However, offline RL methods are susceptible to distributional discrepancies that might arise between the behavior policies used during the meta-train and meta-test phases \cite{yuan2022robust}. In fully offline meta RL scenarios where both the meta-train and meta-test contexts are gathered using unknown behavior policies, the performance of trained RL agents in testing tasks heavily depends on the quality of the meta-test contexts. If the training policy is unbalanced concerning policy quality-for instance, if it only comprises expert data-the learned context encoder may not generalize well to a broader range of meta-test policies. We utilize the \textit{PointRobotGoal} environment to illustrate how the anticipated contextual embeddings and resultant trajectories differ based on the quality of the contextual data, as depicted in Fig.~\ref{fig: quality}. As shown, performance drops dramatically when the context information is pre-collected with policies dissimilar from those used for meta-training.

Motivated by these observations, we aim to tackle the distribution shift problem for OMRL and present an importance sampling method to learn a more robust task context encoder. We adopt the supervised contrastive learning framework \cite{khosla2020supervised} with task indexes as labels. Then, we calculate the ``hardness" of positive and negative samples separately based on their distances to the anchor sample in the embedding space. Finally, we adjust the supervised contrastive loss function by weighing both positive and negative samples according to their ``hardness" values.

Our main contributions include:
\begin{itemize}
    \item
     We analyze the distribution shift problem in learning a task context encoder for OMRL and demonstrate the impact of context quality on the performance of a learned context encoder.
    \item
     Based on our analysis, we propose a novel supervised contrastive objective that adopts hard positive and hard negative sampling to train a more resilient context encoder. We conduct experiments to demonstrate the superior performance of our proposed approach compared to various baseline methods.
    \item
     Additionally, we perform an ablation study to investigate the contribution of the proposed hard sampling strategy for robust context encoder learning. We also analyze the quality of the learned context encoder concerning its uniformity and alignment characteristics.
\end{itemize}
\section{Related Work}
\subsection{Meta Reinforcement Learning}
Meta-learning has proven to be a successful approach by enabling the learning of knowledge about the learning process itself, resulting in improved learning efficiency and exceptional performance across various applications \cite{schmidhuber1987evolutionary, bengio1990learning, hospedales2021meta}. In the context of meta RL, prior research focused on developing a meta-policy and an adaptation approach that work simultaneously to facilitate sample-efficient task adaptation during testing. Optimization-based approaches allow fine-tuning of the meta-policy using only a few shots of data in the meta-test tasks, leveraging gradient descent \cite{finn2017model,mitchell2021offline}. Alternatively, context-based meta RL methods learn a meta-policy conditioned on some task context information, thereby eliminating the need to update networks during the meta-test phase. Duan et al. \cite{duan2016rl} and Wang et al. \cite{wang2016learning} utilized a recurrent neural network to encode the context information, while Rakelly et al. \cite{rakelly2019efficient}, Zintgraf et al. \cite{zintgraf2019varibad}, and Humplik et al. \cite {humplik2019meta} learned a separate network that encodes context information as task-specific latent variables. With fine-tuning operations in the meta-test stage, optimization-based methods perform more robustly on out-of-distribution test tasks. However, this comes at the cost of increased computing resources during meta-testing. On the other hand, context-based methods achieve higher sample efficiency and better asymptotic performance when adapting to in-distribution target tasks, such as reaching different goal positions or running at different target speeds \cite{rakelly2019efficient}.

Recently, various works extended meta RL approaches to offline settings, assuming that agents can only access static datasets of pre-collected experiences in each training environment, rather than the environments themselves, to address offline RL scenarios where online data collection is not feasible \cite{levine2020offline, prudencio2022survey}. Mitchell et al. \cite{mitchell2021offline} utilized optimization-based methods and applied a Model-Agnostic Meta-Learning (MAML) style loss in policy learning.

Regarding context-based methods, Li et al. \cite{li2020focal} disentangled the learning of the encoder and policy. They first learned an informative task context encoder with transition data only and then learned the conditional policy using standard offline RL methods, such as Behavior Regularized Actor Critic (BRAC) \cite{wu2019behavior}. Li et al. \cite{li2021provably} and Yuan et al. \cite{yuan2022robust} employed the objective of contrastive learning for learning task representation and achieved more robust performance in testing scenarios. However, most prior works assumed that the underlying distribution of context data remained unchanged between training and testing and did not explicitly consider the problem of distribution shift. This shift is common in OMRL, for instance, when agents are trained on experience data primarily collected with expert-level policies but need to explore new environments with random or sub-optimal policies.

To tackle the issue of distribution shifts in OMRL, we propose a hard sampling approach to context encoder learning.
\subsection{Contrastive Learning}
Contrastive learning is a prevalent technique for representation learning \cite{oord2018representation, chen2020simple, he2020momentum}. It learns data representations by encouraging similar samples to remain close in the embedding space while pushing away dissimilar ones. In self-supervised learning settings where there is no label information available, the positive pair refers to various augmented views of a single sample, while negative samples refer to views of different samples \cite{chen2020simple}. In supervised learning settings, samples that belong to the same class label are considered positive samples and their label information is embedded into contrastive objectives \cite{khosla2020supervised}.

Several recent works have focused on analyzing the behavior of contrastive learning. Robinson et al. \cite{robinson2020contrastive} emphasized the importance of negative sample distribution and proposed an importance sampling technique to mine hard negative samples. Furthermore, Wang et al. \cite{wang2020understanding} and Wang et al. \cite{wang2021understanding} addressed two essential properties of learned representations that contribute to good performance in downstream tasks: \textit{alignment}, which measures the proximity between similar samples in the embedding space, and \textit{uniformity}, which measures how similar the learned representation distribution is to a uniform distribution. Theories provided by \cite{huang2021towards} reveal that it is crucial to strike a balance between these two critical properties to find good representations.
\section{Preliminaries}
In meta RL, we consider a distribution of tasks $p_{\mathcal{T}}(\cdot)$. Each task is formalized as a Markov decision process (MDP), defined as $\mathcal{T} = \langle\mathcal{S,A}, P, R, \gamma \rangle$, where $\mathcal{S,A}, P, R, \gamma$ denote the state space, action space, transition function, reward function and discount factor, respectively. The same as previous works \cite{rakelly2019efficient, li2020focal}, we assume here that similar tasks share the same state space $S$, action space $A$ and discount factor $\gamma$. The differences among tasks lie in the reward functions (e.g., reaching to different goals) and transition functions (e.g., walking on different terrains). During the meta-train stage, the agents have full access to a set of meta-train tasks $\{\mathcal{T}\}^{\text{train}}$. The objective of meta RL is to train an agent with data from the meta-train task set only, such that the trained agent can quickly adapt itself to unseen target tasks $\{\mathcal{T}\}^{\text{test}}$ with limited data.
\subsection{Context-Based Meta RL}
To enable quick adaptation to target tasks, context-based methods learn both a task context encoder parameterized by $\phi$: $z\sim q_\phi(z|x)$ and a contextual policy parameterized by $\theta$: $a\sim\pi_\theta(a|s, z)$. Specifically, the encoder predicts a latent task embedding by encoding some context information $x$, whereas the contextual policy predicts the optimal action $a$ given the current state $s$ and the encoded embedding $z$. In situations where there is a lack of prior knowledge regarding the target task, the context information $x$ within meta RL may consist of a limited quantity of interactive experience $\{s_t, a_t, s_{t+1}, r_t\}_{t=1,2,...,T}$ that is relevant to the target task.

Specially for OMRL, Li et al. \cite{li2020focal} discovered that training the context encoder separately from the contextual value function and policies can lead to more effective results. Specifically, during the meta-training stage, the agent initially learns the context encoder, which is then frozen before proceeding to train the RL components such as the actor and critic networks. At the meta-test stage, given some task context information $\tau^{\text{te}}$, e.g., one trajectory data collected in testing environment, the agent first samples the task representation $z^{\text{te}}\sim q_\phi(z|\tau^{\text{te}}) $, then deploys the corresponding acting policy $\pi_\phi(a|s, z^{\text{te}})$ into the new environment.
In this work, we also choose to decouple the learning of the context encoder and contextual policy, since this approach has been shown to improve robustness and lead to better overall performance.

By treating the latent task embedding $z$ as an unobserved component of the state, we can frame the context-based meta RL problem as a partially observed MDP (POMDP). Specifically, in this POMDP formulation, the state comprises both the environment state $x$ and the task embedding $z$. Since the task remains fixed within a given trajectory, the task embedding $z$ also remains constant over that period. If we assume that the task context encoder performs well, then the learning of the acting policy $\pi_\phi(a|s,z)$ and associated value functions $V(s,z), Q(s,a,z)$ can be addressed using any standard RL algorithms.
\subsection{Offline RL} The offline RL approach assumes that the agent can only access a static data buffer of trajectories that were collected using an unknown behavior policy $\pi_b$. One issue with this method is that conventional techniques for learning an optimal Q function may require querying unseen actions, which can cause estimation errors and an unstable learning process. To tackle this challenge, Dorfman et al. \cite{dorfman2021offline} introduced implicit Q-learning (IQL), which employs expectile regressing to learn the optimal value function $V_\psi(s)$ and Q function $Q_\theta(s,a)$. The corresponding loss functions are given as follows:
\begin{equation}
\label{eq: vf}
        \mathcal{L}_V(\psi) = \mathbb{E}_{(s,a)\sim \mathcal{D}} \big[ L_2^{\tau}(Q_{\hat{\theta}}(s,a) - V_\psi(s)) \big],
\end{equation}
\begin{equation}
\label{eq: qf}
    \mathcal{L}_Q(\theta) = \mathbb{E}_{(s,a,s') \sim D} [(r(s,a) + \gamma Q_{\hat{\theta}}(s',a') - Q_{\theta}(s,a))^2],
\end{equation}
where $\mathcal{D}$ is the offline dataset, $\theta$, $\hat{\theta}$, and $\psi$ parameterize the Q network, target Q network, and the value network, respectively. $L_2^{\tau}(x) = | \tau - \mathbbm{1}(x<0) |x^2$ is the expectile regression function.

To learn policies, advantage-weighted regression is utilized to extract a policy that is parameterized by  from the estimated optimal Q function \cite{peng2019advantage}. The objective is
\begin{equation}
    L_{\pi}(\phi) = \mathbb{E}_{(s,a)\sim \mathcal{D}}\big[\exp(\beta(Q_{\hat{\theta}}(s,a) - V_{\psi}(s))) \log \pi_\phi(a|s)  \big].
\end{equation}

Given its robustness and effectiveness in the context of offline RL, we adopt IQL as our method of choice for downstream policy learning in this work.

\section{Learning Task Representation}
Offline RL poses the challenge of learning from data that was collected using unknown policies. As a result, distribution shifts between training and testing data due to performance differences in these policies can significantly degrade testing performance \cite{levine2020offline}. Similarly, imbalanced training data during task representation learning may lead to poor generalization when presented with contexts from different behavior policies. For instance, if offline data is primarily gathered using near-optimal policies, the learned context encoder could fail to identify low-quality context data, such as trajectories generated by worse policies.

To overcome this challenge, we propose a contrastive learning and importance sampling-based method for task representation learning from imbalanced offline training sets. By assuming that contexts of various qualities will be encountered uniformly during the meta-test stage, we weigh both positive and negative samples in the supervised contrastive objective according to their ``hardness", which is measured by the distance between samples in the embedding space.
\subsection{Representation Learning Framework}
\label{sec: framework}
Consider the common meta RL scenario in which no prior information about tasks is available, and the agent must infer about the task based on trajectory data $\tau~=~\{s_t, a_t, s_{t+1}, r_t\}_{t=0,1,...,T}$. To process a batch of data, we first apply data augmentation to generate two distinct views of the batch. In the case of trajectory data, we use both a transition encoding module and a trajectory aggregation module to produce normalized embeddings. During the training phase, we also employ a projection module to generate lower-dimensional projections, upon which contrastive losses are computed. It is worth noting that the projection module will be discarded after the representation learning stage.

To summarize, the key components of our representation learning framework are as follows:

\keypoint{Data augmentation} For every input trajectory $\tau$, we create two distinct views of the data, with each view containing a subset of the information found in the original sample. Similar to how images are cropped in computer vision, we randomly select two different segments $\tau^1, \tau^2$ as the two views of the trajectory $\tau$.

\keypoint{Encoding network} The encoding network is responsible for mapping the input trajectories to latent representation vectors $\mathbf{z}~=~Enc(\tau)~\in~\mathbb{R}^{D_{E}}$, where $D_E$ represents the dimension of the embedding space. The encoding network comprises two components: First, a transition encoder that maps each transition to a transition embedding vector $\mathbf{v}~=~f(s,a,s',r)~\in~\mathbb{R}^{D_{T}}$; second, an aggregator network that collects information from all the transitions and produces the latent representation $\mathbf{z}~=~g(\{\mathbf{v}_t\})\in~\mathbb{R}^{D_{E}}$. We normalize the embeddings $\textbf{v}$ and $\textbf{z}$ such that they lie on the unit hypersphere in $\mathbb{R}^{D_{E}}$ and $\mathbb{R}^{D_{T}}$, respectively.

\keypoint{Projection network}
The projection network is responsible for mapping the context embedding $\textbf{z}$ to the final output vector $\textbf{w}$, where $w~=~Proj(z)\in \mathbb{R}^{D_P}$, and is used in computing the distance between a pair of data samples. We normalize the projected output $w$ to lie on a unit hypersphere. Following prior work \cite{chen2020simple, khosla2020supervised}, we discard the projection head when performing downstream tasks such as learning the contextual policy within the context of OMRL.
\subsection{Contrastive Objective}
Consider a set of training tasks $\{\mathcal{T}^k\}$ and the corresponding offline datasets $\{\mathcal{B}^k\}$. Each dataset $\mathcal{B}^k$ comprises trajectories $\{\tau_n^k\}$ collected offline in task $\mathcal{T}^k$ using an unknown behavior policy, where $k\in I \equiv {\{1,...,K \}}$ denotes the index of an task, and $n\in I \equiv \{1,...,N\}$ denotes the index of an arbitrary trajectory sample. The objective of representation learning is to encode the trajectory data $\tau$ to a latent representation vector $z_{\tau}$, such that trajectories from the same task are similar to each other in the embedding space. To achieve this, we first apply data augmentation to generate two different views. Let $\mathcal{D}^i \triangleq \{A^i(\tau) | \tau\in \bigcup\{\mathcal{B}^k\}\}_{i=1,2}$ denote the $i$th view of the trajectories drawn from all tasks, and let $\mathcal{D} \triangleq [\mathcal{D}^1, \mathcal{D}^2]$ denote the collection of two views.

\keypoint{Supervised contrastive learning (SCL)} Since the task labels are available in our OMRL setup, we leverage supervised contrastive learning to incorporate the information of task labels into the dataset \cite{khosla2020supervised}. Given an anchor sample $\tau_q = A^1(\tau)$, we consider the other view $A^2(\tau)$. Pairs of samples with the same task label in these two views are treated as positive samples, while those associated with different task labels are treated as negative samples. Following the approach taken in \cite{oord2018representation}, we adopt the InfoNCE loss for this purpose, resulting in the following objective function
\begin{align}
    \mathcal{L}^{\text{SCL}} = \sum_{\tau_q\in \mathcal{D}} \frac{-1}{|P(\tau_q)|}\sum_{\tau_p\in P(\tau_q)} \log\frac{e^{z_{\tau_q} \cdot z_{\tau_p} / \beta}}{ \sum_{a\in\mathcal{D}(\tau_q)} e^{z_{\tau_q}\cdot z_a/\beta}  },
    \label{eq: supcon}
\end{align}
where $\tau_q$ denotes the anchor sample, $ P(\tau_q) $ the set of corresponding positive samples, $\mathcal{D}(\tau) \equiv D\backslash \{\tau_q\}$ the collection of views excluding the anchor sample $\tau_q$.

\keypoint{Hard negative sampling} To automatically select more informative negative samples, Robinson et al. \cite{robinson2020contrastive} proposed hard negative sampling. This technique draws inspiration from importance sampling methods, where the optimizer places greater emphasis on challenging negative samples. These are samples that are located in close proximity in the embedding space but are associated with different tasks. In this work, we extend the hard negative sampling method to the supervised contrastive learning framework by defining the loss function as follows:
\begin{align}
    \mathcal{L}^{\text{HG}} = \sum_{\tau_q\in \mathcal{D}} \frac{-1}{|P(\tau_q)|}\sum_{\tau_p\in P(\tau_q)} \log\frac{e^{z_{\tau_q} \cdot z_{\tau_p} / \beta}}{ Z^{\text{pos}} + Z^{\text{neg}}}, & \label{eq: hard_neg}\\
    Z^{\text{pos}} = \sum_{a\in\mathcal{P}(\tau_q)} \exp(z_{\tau_q} \cdot z_{a}/\beta), & \\
    Z^{\text{neg}} = \sum_{a\in\mathcal{N}(\tau_q)} \omega^{\text{neg}}_{a}~\exp(z_{\tau_q}\cdot~z_{a}/\beta), &
\end{align}
where $Z^{\text{pos}}$ is computed over all positive samples $\mathcal{P}(\tau_q)$, and $Z^{\text{neg}}$ over all negative ones $\mathcal{N}(\tau_q)$. Here, $\omega^{\text{neg}}_{a}$
measures the ``hardness'' of a negative sample $a$, namely, $\omega^{\text{neg}}_{a} \triangleq {\exp{(z_{\tau_q} \cdot z_{a})}} / \sum_{a'\in \mathcal{N}(\tau_q)}{\exp{(z_{\tau_q} \cdot z_{a'})}}$.

\keypoint{Hard positive sampling} To further address the issue of imbalanced data in representation learning for OMRL, we propose re-weighting the positive samples in Eqn.~\ref{eq: hard_neg} based on their ``hardness''. Specifically, we assign higher weights to positive samples that are further away from the anchor sample. As a result, the loss function is formulated as follows:
\begin{align}
    &\mathcal{L}^{\text{HP+HG}} = \sum_{\tau_q\in \mathcal{D}} \frac{-1}{|P(\tau_q)|}\sum_{\tau_p\in P(\tau_q)} \log\frac{\omega^{\text{pos}}   _{\tau_p}\cdot e^{z_{\tau_q} \cdot z_{\tau_p} / \beta}}{ \hat{Z}^{\text{pos}} + Z^{\text{neg}}}
    \label{eq: hard_pos},
\end{align}
where $\hat{Z}^{\text{pos}} = \sum_{a\in\mathcal{P}(\tau_q)} \omega^{\text{pos}}_{a} \exp(z_\tau \cdot z_a/\beta)$ is the re-weighted sum over positive samples, and $Z^{\text{neg}}$ remains unchanged. Here,
$\omega^{\text{pos}}_{a}$ measures the ``hardness'' for a positive sample $\tau_p$, namely, $\omega^{\text{pos}}_{a}= \exp(-z_{\tau_q} \cdot z_{a}) / \sum_{a'\in\mathcal{P}(\tau_q)}\exp(-z_{\tau_q} \cdot z_{a'})$. Note that the positive weights are calculated based on the negation of dot products. Therefore, positive samples that are farther away from the anchor sample will receive more significant weights in the gradient computation.
\section{Experiments}
\label{sec: experiments}
In the experiments, our main objective is to assess whether our proposed method can develop a more robust context encoder, particularly when the training data is imbalanced. Our initial focus is on highlighting the issue of context distribution shift in task representation learning for OMRL. We then proceed to compare our technique with baseline methods, specifically in terms of their ability to handle contexts of varying qualities. Finally, we conduct two additional experiments to gain further insight into our approach. These involve reviewing the sampling strategy of our method and analyzing the uniformity and alignment characteristics of the trained encoders.
\subsection{Experimental Setup}
We utilize a total of five simulated continuous control environments, comprised of three environments featuring varying goal conditions (reward functions) and two with dynamic variations (transition functions). These specific tasks have been previously employed in studies such as \cite{rakelly2019efficient,yuan2022robust,li2020focal}.

\keypoint{Environments with changing reward functions}

\textit{PointRobotGoal}: the agent is responsible for controlling a robot to navigate towards various goal positions located on a unit circle. The reward is determined by the Euclidean distance between the robot and the designated goal position;

\textit{AntDir}: the agent is tasked with controlling a simulated ant robot to move in a variety of two-dimensional directions. The reward is determined by the speed projected on the desired direction of movement;

\textit{CheetahVel}: the agent is responsible for controlling a simulated half-cheetah robot to run at varying target speeds. The reward is calculated as the difference between the actual speed of the robot and the designated target speed.

\keypoint{Environments with changing transition functions}
The \textit{WalkerParams} and \textit{HopperParams} environments are both locomotion tasks simulated using MuJoCo. The goal of each task is to train the agent to propel a walker or hopper robot forward as quickly as possible. In each scenario, certain physical parameters such as the robot's mass and friction coefficient are randomized.

For every environment, a total of 30 tasks are uniformly sampled to create the training set of tasks. Additionally, 10 tasks are sampled from the same distribution as the target task set. To generate the offline dataset, we utilize soft actor-critic (SAC) \cite{haarnoja2018soft} to train each individual task separately. The data captured during this process is stored as the offline dataset in the form of a replay buffer. More details about the testing environment distribution and the offline dataset are referred to the Appendix Section.
\subsection{The Distribution Shift Problem}
Our experiments revealed the previously mentioned distribution shift issue. In the \textit{PointRobotGoal} environment, we visually observed the context embeddings and downstream performance outcomes when provided with contexts of varying qualities. This is illustrated in Fig.~\ref{fig: quality}. To generate testing context buffers of varying qualities, we began by ranking trajectory samples from the offline context buffer for each task based on their accumulated returns. Next, the entire buffer was uniformly split into 10 smaller buffers. The context buffer of low quality was comprised of samples that had a performance level within the lowest 10th percentile. Context buffers of medium (and high) quality were made up of samples with a performance level ranging from the 10th to 20th percentile (and the top 10th percentile), respectively. The encoder and policy networks were meta-trained according to the SCL objective (Eqn.~\ref{eq: supcon}). During the meta-test stage, one trajectory was sampled from the buffer as context information each time, which allowed us to produce one outcome trajectory with the inferred task representation and contextual policy.

\begin{figure}[t]
\centering
\begin{subfigure}[b]{0.40\textwidth}
\includegraphics[width=\textwidth]{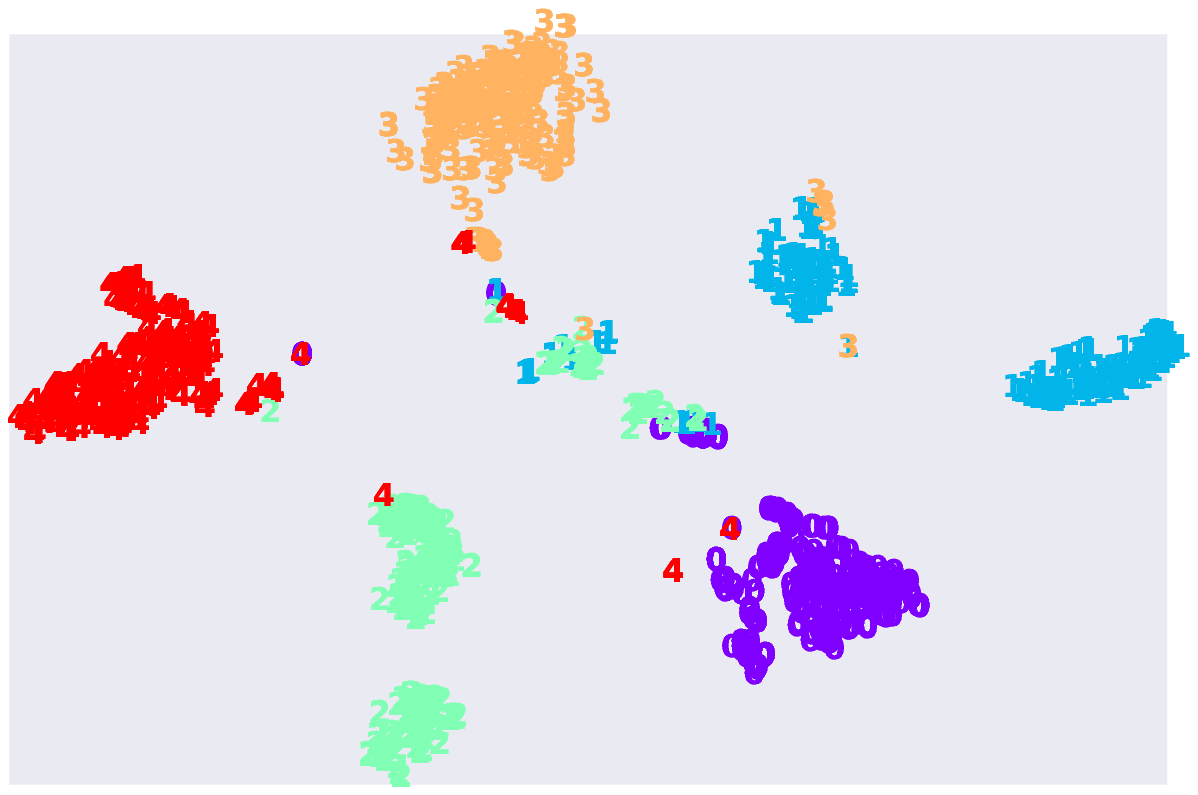}
\caption{t-SNE visualization of context embeddings, trained with hard sampling}
\end{subfigure}
\begin{subfigure}[b]{0.40\textwidth}
\includegraphics[width=\textwidth]{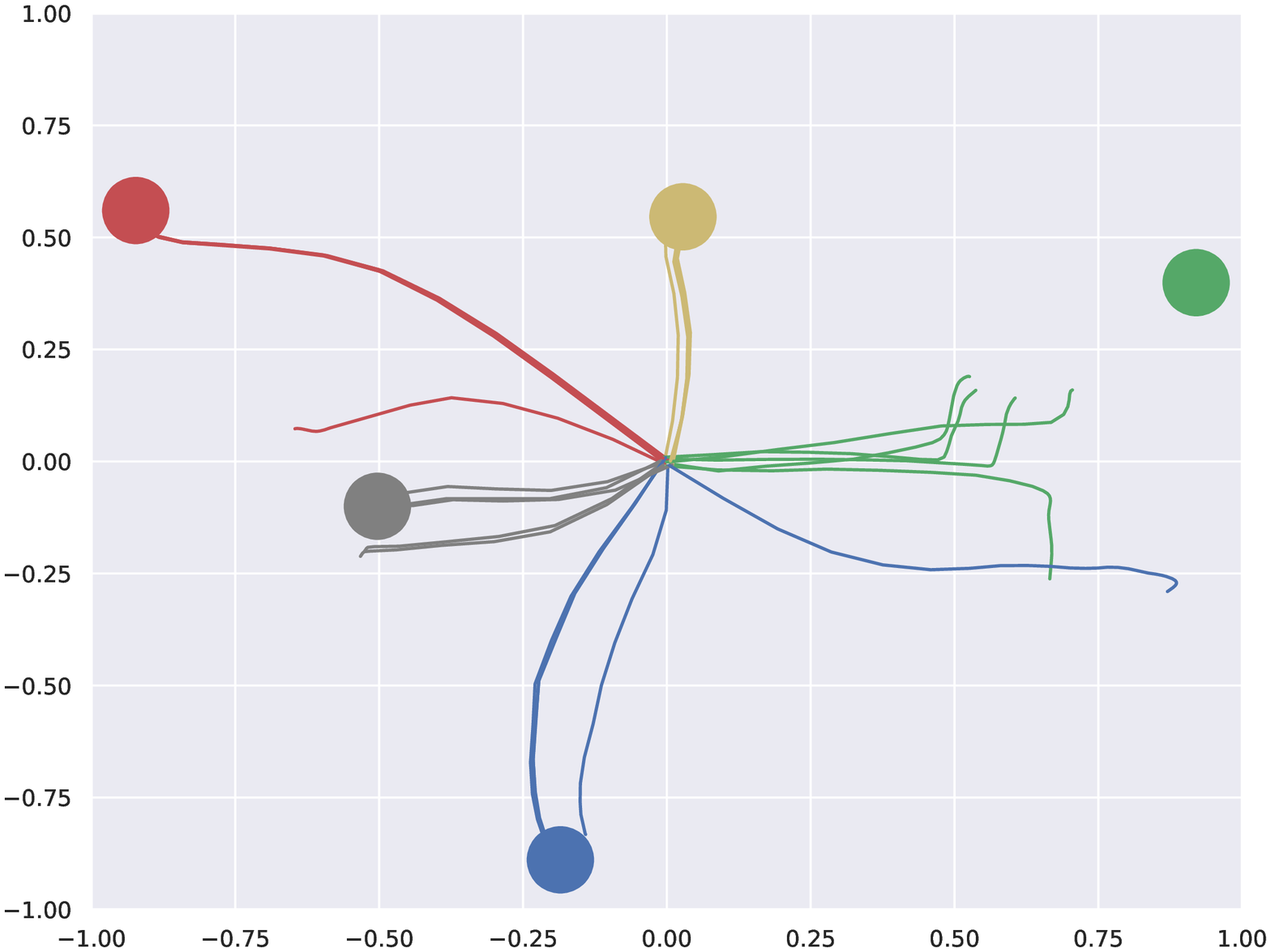}
\caption{trajectories generated by policy, trained with hard sampling}
\end{subfigure}
\begin{subfigure}[b]{0.40\textwidth}
\includegraphics[width=\textwidth]{figs/low_quality_tsne.eps}
\caption{t-SNE visualization of context embeddings, trained without hard sampling}
\end{subfigure}
\begin{subfigure}[b]{0.40\textwidth}
\includegraphics[width=\textwidth]{figs/low_quality_traj.eps}
\caption{trajectories generated by policy, trained without hard sampling}
\end{subfigure}
\caption{t-SNE visualization and demonstrative trajectories given contexts of low-qualities in the \textit{PointRobotGoal} environment, comparing between training with hard sampling (a,b) and without hard sampling (c,d). The results show that hard sampling strategy helps with making task contexts more separable in the embedding space and consequently generating better trajectories against low-quality contexts.}
\label{fig: tSNE}
\end{figure}

In Fig.~\ref{fig: quality}, we have provided t-SNE visualizations of the resulting embeddings along with exemplary outcome trajectories. As demonstrated in the figure, the learned context encoder is capable of separating high-quality contexts (located in the right-hand column) in the embedding space and achieving the target goal. However, as the quality of the context information decreases (as observed in the left and middle columns), the performance drastically deteriorates. When incorporating context information that was obtained using dissimilar policies from those utilized during meta-training, the encoder trained according to SCL fails to cluster the context embeddings properly, which results in inferior testing performance outcomes. In comparison to SCL, our proposed method (Fig.\ref{fig: tSNE}) greatly improves the separability of task contexts in the embedding space, thereby producing better testing trajectories against low-quality contexts.
\subsection{Task Adaptation Performance}
\label{sec: low-qual}

To evaluate the effectiveness of our hard sampling approach to OMRL, we compared it against the following baseline methods:

\keypoint{Offline PEARL} A natural baseline for OMRL is to extend off-policy meta RL approaches to the offline setup. In this regard, we have considered the offline variant of PEARL as a baseline method \cite{rakelly2019efficient}. In offline PEARL, the context encoder is jointly trained with the contextual value functions and contextual policy. The Mean Squared Error (MSE) loss in estimating the value function is utilized to update the context encoder.

\keypoint{FOCAL} Li et al. \cite{li2020focal} employed metric learning to train the context encoder. The aim is to bring positive pairs of samples together while pushing negative pairs apart. Positive and negative pairs are defined as transition samples from the same and different tasks, respectively. Another variant of FOCAL, called FOCAL++ \cite{li2021provably}, has also been considered. The primary differences include that FOCAL++ replaces the objective function with a contrastive objective with momentum \cite{he2020momentum} and employs attention blocks to encode the trajectories instead of transitions.

\keypoint{CORRO} Yuan et al. \cite{yuan2022robust} generated synthetic transition samples as negative samples and trained the encoder network using InfoNCE loss \cite{oord2018representation}. In our experiments, we utilize reward randomization to generate synthetic transitions.

Note that for analyzing the impact of task representation learning, we have employed IQL \cite{kostrikov2021offline} as the underlying offline RL method for all baselines. Although the original implementations of CORRO and PEARL employ SAC \cite{haarnoja2018soft} and FOCAL uses BRAC \cite{wu2019behavior}, respectively, we have conducted two experiments where the agents are trained with the original offline RL algorithms and with IQL, respectively. We have found that IQL performs better in all cases, and therefore only report results trained with IQL.

In the following experiments, we have evaluated the test performance of meta-trained policies in scenarios where distribution shifts exist between the meta-train and meta-test stages. Similar to the previous experiment, the low-quality context buffer comprises trajectory data with accumulated returns within the lowest 10th percentile. During testing, we randomly select one trajectory from the context buffer each time and test the agent in the corresponding task. The performance outcomes are measured by the accumulated return and are averaged over all tasks in the target task set.

Table~\ref{tab: main} summarizes the test results in three separate environments, where all policies are tested using low-quality contexts. As observed in the experiments, FOCAL and CORRO encounter challenges when trying to generalize to low-quality contexts during testing. We believe that this is largely due to the fact that the training buffers mainly consist of high-quality data, which causes the baseline methods to fail in capturing the information contained in low-quality context data. In comparison to these baselines, our hard sampling strategy helps to achieve better overall performance in these environments.

In the other two environments, \textit{CheetahVel} and \textit{HopperParams}, our proposed methods, hard sampling performs competitively when compared with CORRO. The testing performances of our $\mathcal{L}^{\text{HP+HG}}$ are $-67.5$ and $182.3$, respectively, whereas those of CORRO are $-65.3$ and $180.7$. This can be attributed to the fact that in these environments, the contexts are less diverse due to the nature of the environment. For instance, consider the \textit{CheetahVel} environment where different tasks are defined as running at different speeds. The robot agent is expected to move forward across all tasks. Consequently, the underlying distribution of the trajectory data is less diverse compared to the other environments (such as \textit{AntDir}). Thus, the baseline methods suffer fewer distribution shift issues in these environments, and the advantage of using hard sampling strategies disappears. For more details on the experimental results, see the Appendix. 

\begin{table*}[t]
    \centering
    \begin{small}
    \begin{tabular}{c c c H c H}
    \hline
        Environment & \textit{PointRobotGoal} & \textit{AntDir} & \textit{CheetahVel} & \textit{WalkerParams} & \textit{HopperParams}\\ 
        \hline
        FOCAL & $-80.4\pm 20.4$& $247.8\pm 54.7$ & $-107.2\pm 27.9$ & $154.8\pm64.3$& $107.3\pm39.8$ \\
        FOCAL++ & $-63.9\pm10.9$ & $289.8 \pm 67.2$  &  $-79.3\pm 29.5$ & $180.7\pm 73.2$& $153.4\pm49.2$ \\
        CORRO & $-52.7\pm14.2$ & $313.0\pm 74.3$ &  $\mathbf{-65.3\pm17.8}$ & $259.0\pm36.4$& $180.7\pm55.7$ \\ 
        Offline PEARL & $-129.4\pm15.7$ & $248.3\pm74.8$  & $-145.1\pm43.8$ & $193.8\pm45.8$& $\mathbf{264.0\pm28.4}$\\
        $\mathcal{L}^{\text{HP+HG}}$ & $\mathbf{-44.9\pm3.5}$ & $\mathbf{352.1\pm42.8}$  & $\mathbf{-67.5\pm14.8}$ & $\mathbf{264.8\pm41.4}$ & $182.3\pm43.5$\\
        \hline
        $\mathcal{L}^{\text{SCL}}$ & $-60.6\pm9.8$ & $309.5\pm63.4$ & $-69.3\pm25.3$ & $160.4\pm58.2$ & $114.2\pm32.7$\\
        $\mathcal{L}^{\text{HG}}$ & $-49.8\pm11.9$ & $337.0\pm48.9$ & $-72.1\pm19.8$ & $247.2\pm33.8$ & $168.2\pm38.6$\\
        $\mathcal{L}^{\text{HP}}$ & $-55.4\pm13.2$ & $328.9\pm43.0$ & $-69.1\pm21.4$ & $183.9\pm42.8$ & $127.1\pm39.0$ \\
        \hline
        
    \end{tabular}
    \end{small}
    \caption{The comparisons of testing performances between our proposed method against baseline methods (top half) and different sampling strategies (bottom half). All learned context encoders are tested with low-quality contexts in target tasks. The \textbf{bold} numbers highlight the best performances over all compared methods. All results are averaged over 5 random seeds.} 
    \label{tab: main}
\end{table*}

\subsection{Further Analysis}

\keypoint{Sampling strategies} The hard sampling strategy serves as a crucial component in our proposed method to address the data imbalance challenge in OMRL. To illustrate how hard positive and hard negative sampling impact the testing performance of the learned policies, we have conducted experiments.

Using the same evaluation protocol, we have compared various hard sampling strategies, including SCL (with no hard sampling), HG (with only hard negative sampling), and HP (with only hard positive sampling where the weights of all negative samples are equal, i.e., $Z^{\text{neg}} = \sum_{a\in\mathcal{N}(\tau_q)} \exp(z_{\tau_q}\cdot~z_{a}/\beta)$ and HG+HP. The testing performances are presented in Table~\ref{tab: main}. As illustrated, utilizing $\mathcal{L}^{\text{HG+HP}}$ leads to the best performance outcomes. Additionally, using HP or HG individually results in better performance when compared to the SCL baseline when tested with low-quality contexts. This suggests that both hard positive and hard negative sampling have a positive impact on learning robust context encoders.

\begin{figure}[t]
    \centering
    \begin{subfigure}[b]{0.40\textwidth}
    \includegraphics[width=\textwidth]{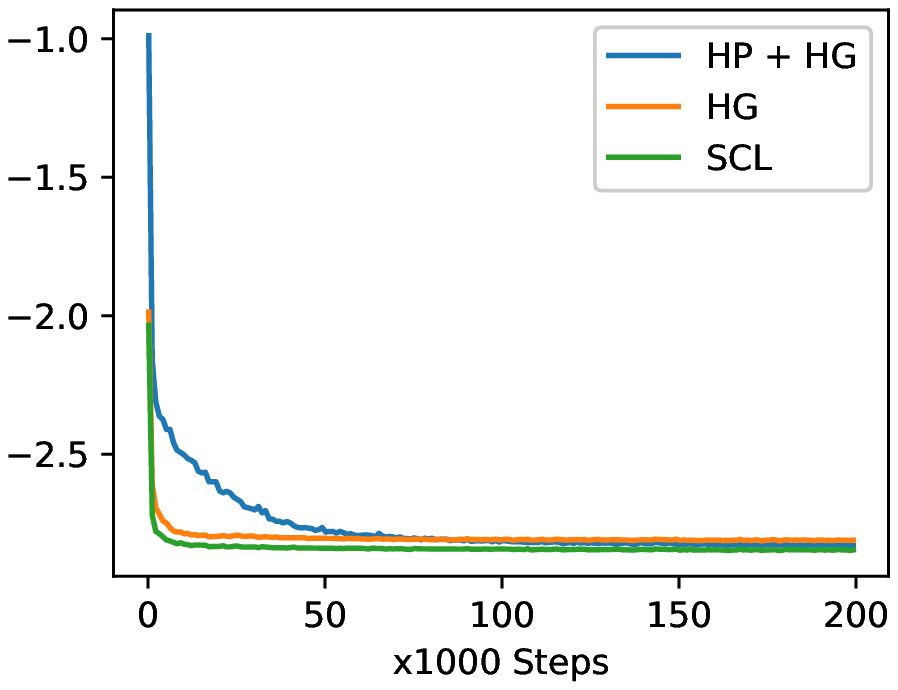}
    \caption{Uniformity}
    \end{subfigure}
    \begin{subfigure}[b]{0.40\textwidth}
    \includegraphics[width=\textwidth]{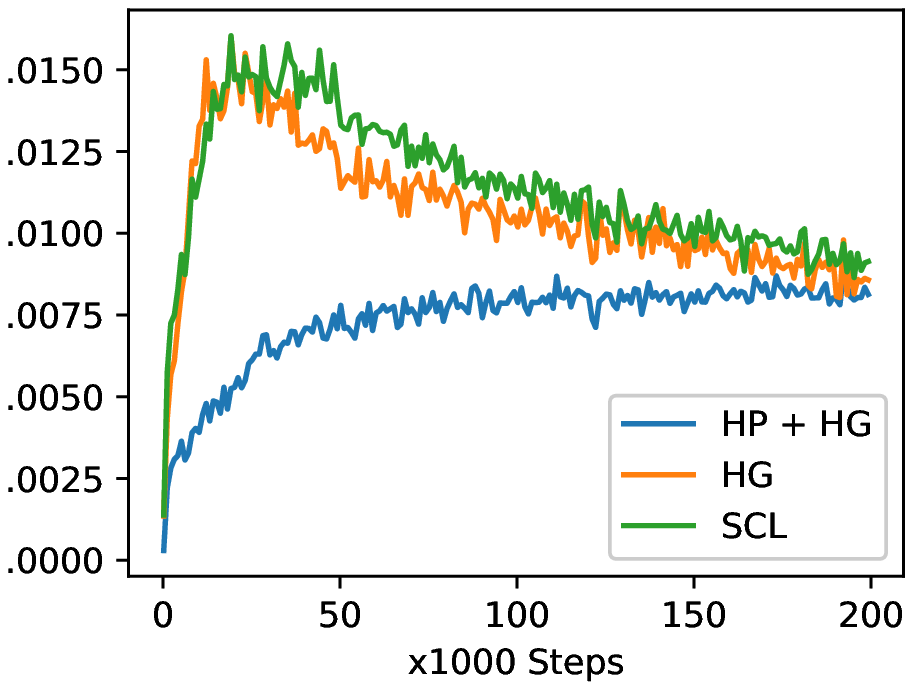}
    \caption{Alignment}
    \end{subfigure}
    \caption{The \textit{uniformity} and \textit{alignment} properties of the embedded contexts in \textit{AntDir} environment given different training strategies. }
    \label{fig: uniform}
\end{figure}

\keypoint{Uniformity and alignment} \textit{Uniformity} and \textit{alignment} are considered to be two vital properties of good representations in contrastive learning \cite{wang2020understanding,wang2021understanding,huang2021towards}. \textit{Uniformity} pertains to how uniformly the samples are spread in the representation space, while \textit{alignment} refers to the closeness between semantically similar samples. Low uniformity may indicate a possible collapse of the model, whereas low alignment implies the loss of semantic information \cite{wang2021understanding}. \cite{wang2020understanding} have found that hard negative mining methods aid in learning embeddings with both low alignment and low uniformity. Ideally, we aim to achieve embeddings with both low alignment loss and low uniformity loss. Formally, given a batch of samples, these two properties can be quantified as follows:
\begin{equation}
\label{eq: uniform}
    \mathcal{L}_{uniformity} = \log \mathbb{E}_{\tau, \tau' \sim \mathcal{D}} \exp(-t|| z_\tau - z_\tau'||^2),
\end{equation}
\begin{equation}
\label{eq: align}
    \mathcal{L}_{alignment} = \mathbb{E}_{\tau \sim \mathcal{D}} \mathbb{E}_{\tau_1,\tau_2 \sim A(\tau)} || z_{\tau_1} - z_{\tau_2} ||^2,
\end{equation}
where $\tau_1, \tau_2$ are a pair of samples augmented from the same sample $\tau$.

We have demonstrated the uniformity and alignment properties of hard sampling compared to other variants in the \textit{AntDir} task. As shown in Fig.~\ref{fig: uniform}, utilizing hard sampling enables the agent to attain a lower alignment loss while the uniformity loss remains comparable across all runs.
\section{Conclusions}
\label{sec: conclusions}
In this paper, we have highlighted that the context distribution shift problem is likely to occur during the task representation learning phase of an offline meta-reinforcement learning (OMRL) process. To address this issue, we have proposed a novel technique that combines the hard sampling strategy with the idea of supervised contrastive learning in the context of OMRL. Our experimental results on several continuous control tasks have demonstrated that when there are context distribution shifts, utilizing our approach can lead to more robust context encoders and significantly improved test performance in terms of accumulated returns, compared to baseline methods. We have open-sourced our code at \url{https://github.com/ZJLAB-AMMI/HS-OMRL} to facilitate future research in this direction towards robust OMRL.
\section*{Acknowledgment}
This work was supported by Exploratory Research Project (No.2022RC0AN02) of Zhejiang Lab.
\bibliographystyle{splncs04}
\bibliography{references}
\section*{Appendix}
\subsection*{Environments and Offline Dataset}
\begin{figure}[h]
\centering
\begin{subfigure}[b]{0.31\textwidth}
\includegraphics[width=\textwidth]{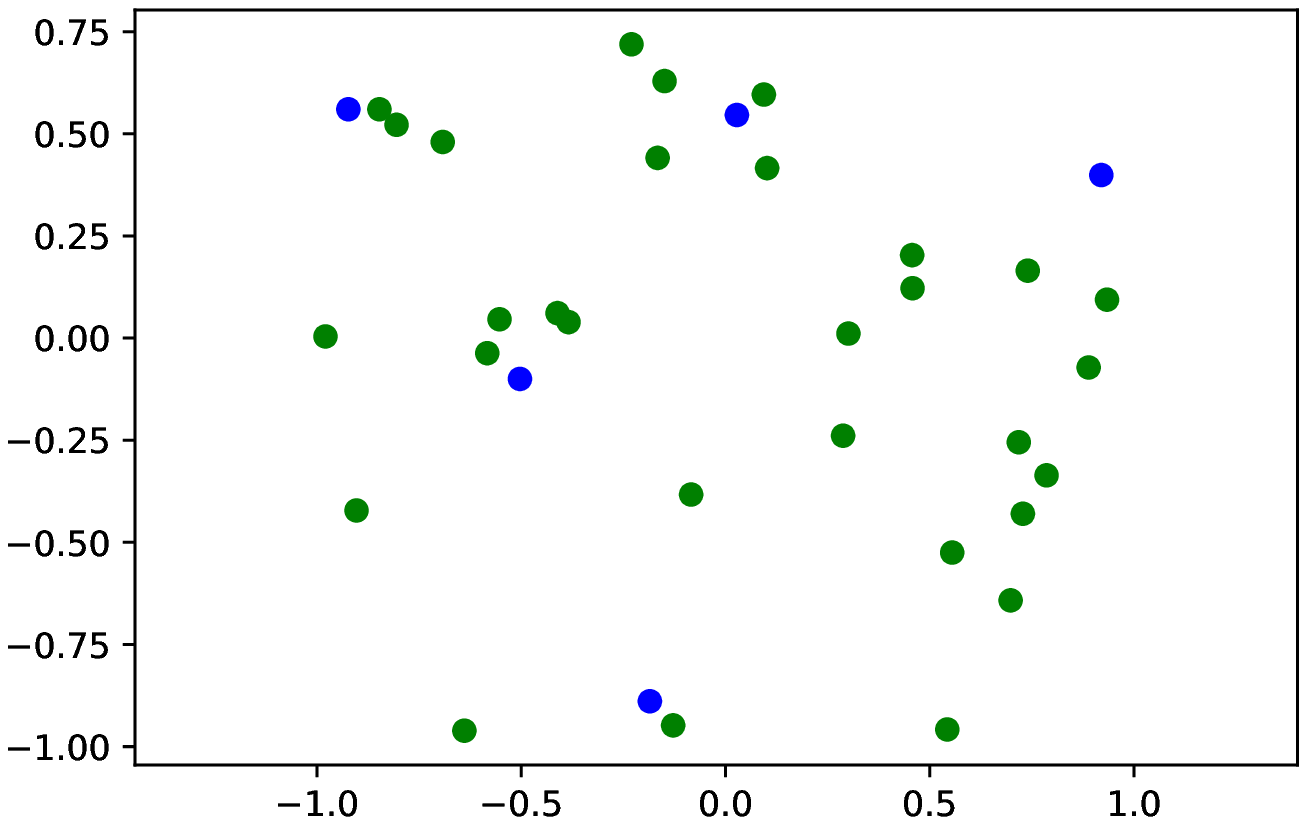}
\caption{PointRobotGoal}
\end{subfigure}
\begin{subfigure}[b]{0.31\textwidth}
\includegraphics[width=\textwidth]{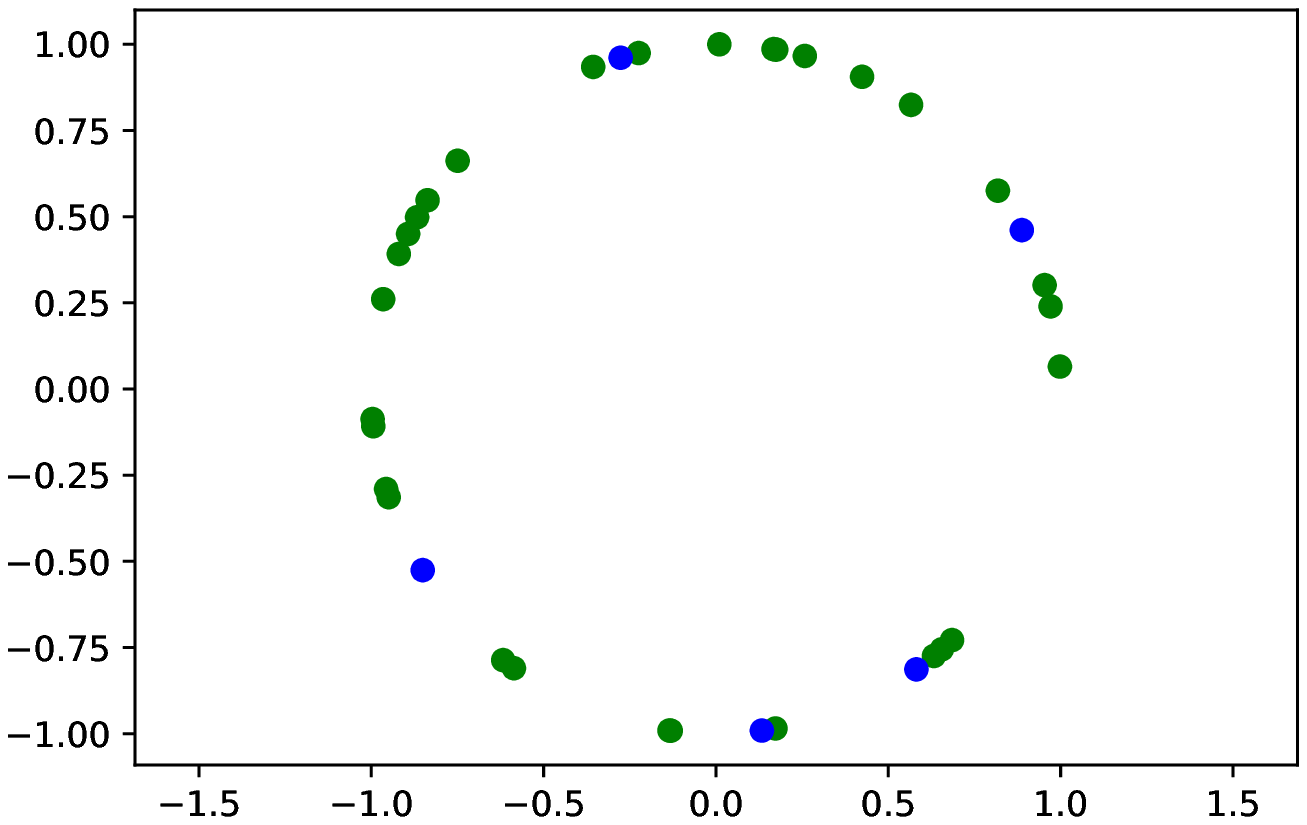}
\caption{AntDir}
\end{subfigure}
\begin{subfigure}[b]{0.31\textwidth}
\includegraphics[width=\textwidth]{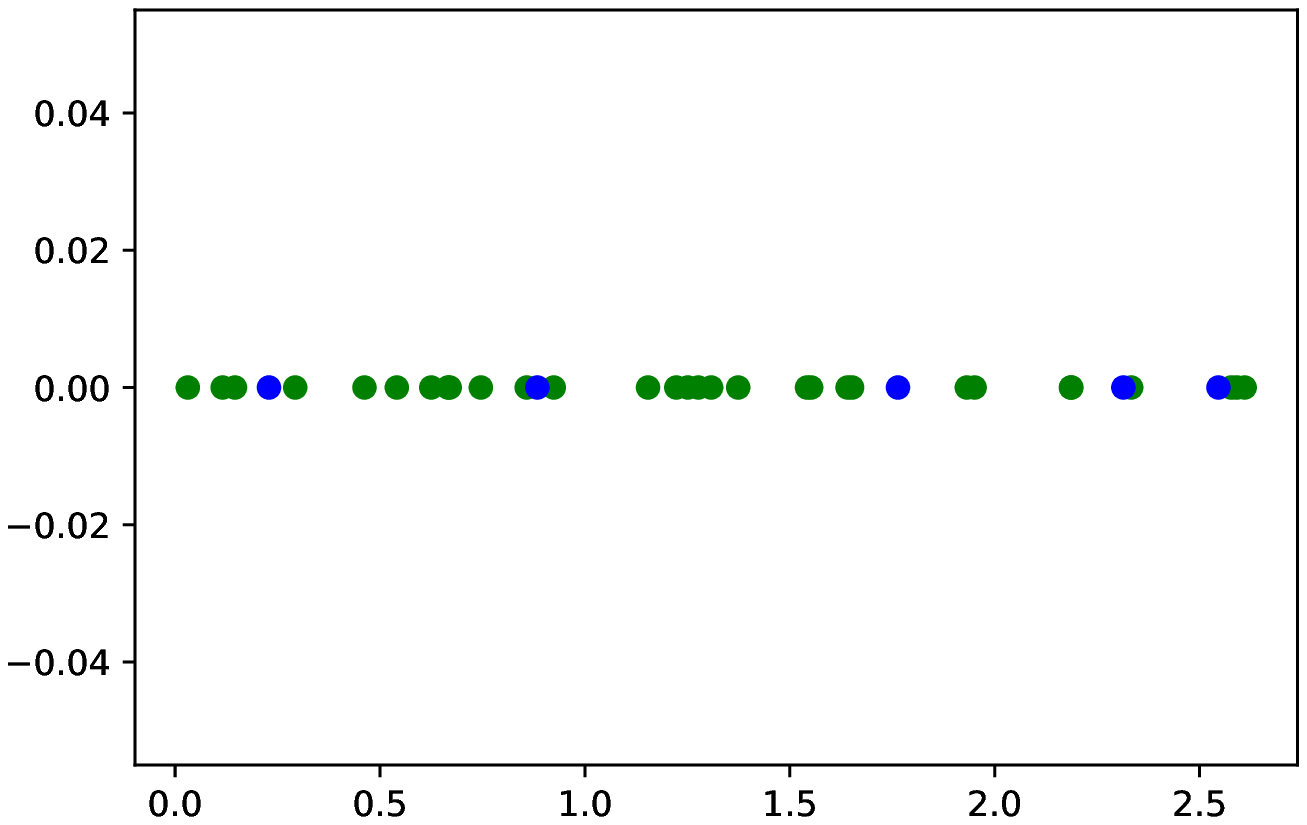}
\caption{CheetahVel}
\end{subfigure}
\caption{Different tasks for each environments. Meta-train tasks are demonstrated with green dots, and meta-test tasks are demonstrated with blue dots.}
\label{fig: goal}
\end{figure}

\begin{figure}[h]
\centering
\begin{subfigure}[b]{0.31\textwidth}
\includegraphics[width=\textwidth]{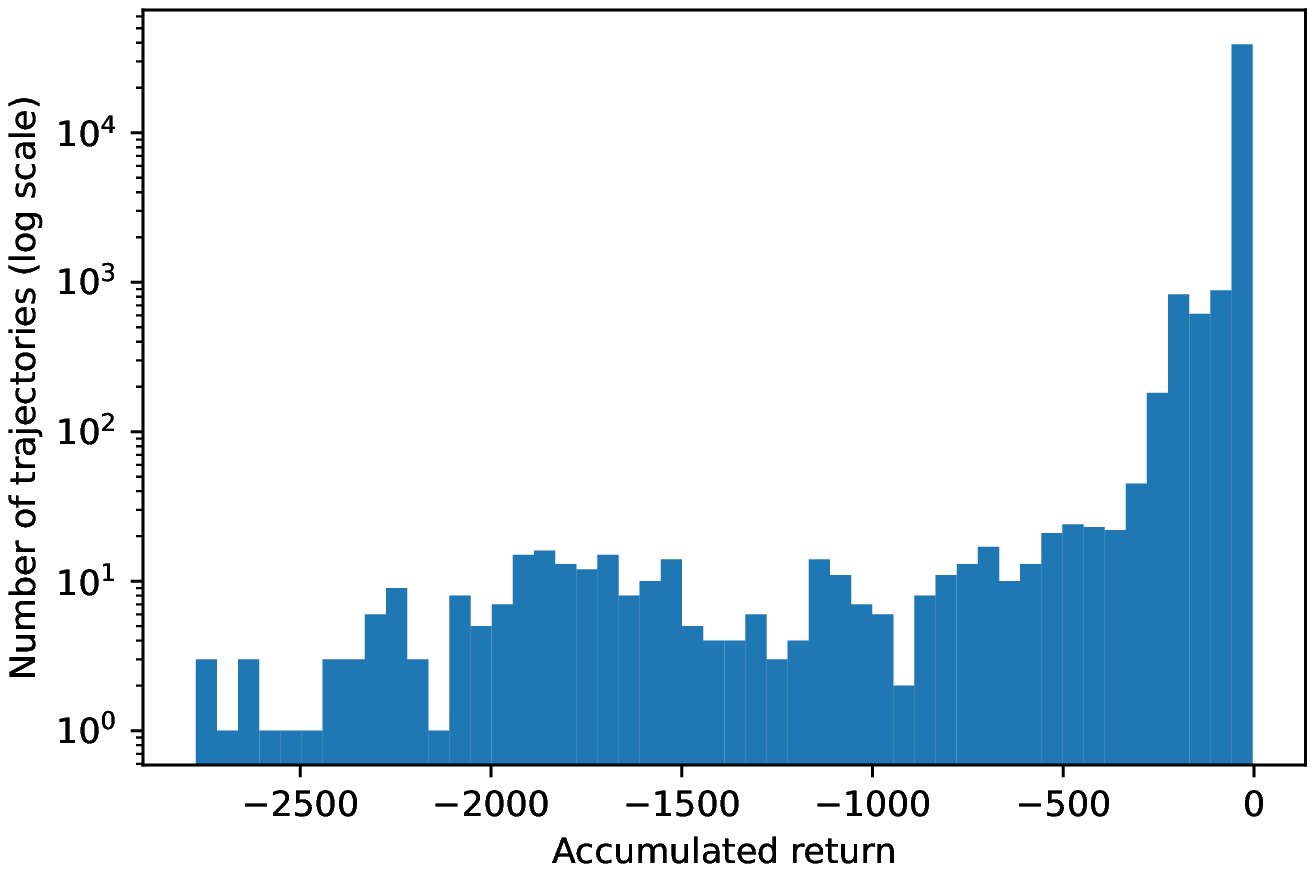}
\caption{PointRobotGoal}
\end{subfigure}
\begin{subfigure}[b]{0.31\textwidth}
\includegraphics[width=\textwidth]{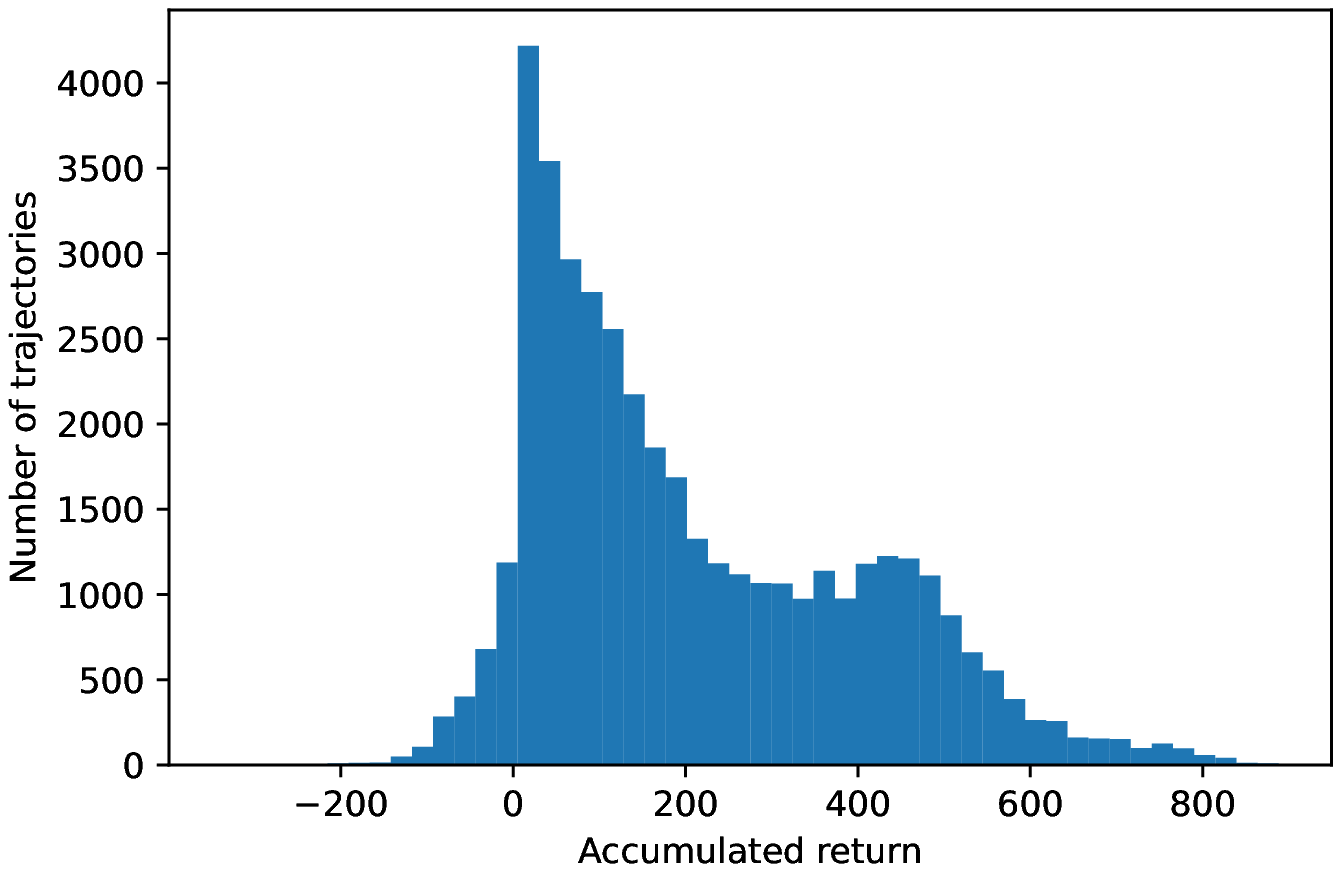}
\caption{AntDir}
\end{subfigure}
\begin{subfigure}[b]{0.31\textwidth}
\includegraphics[width=\textwidth]{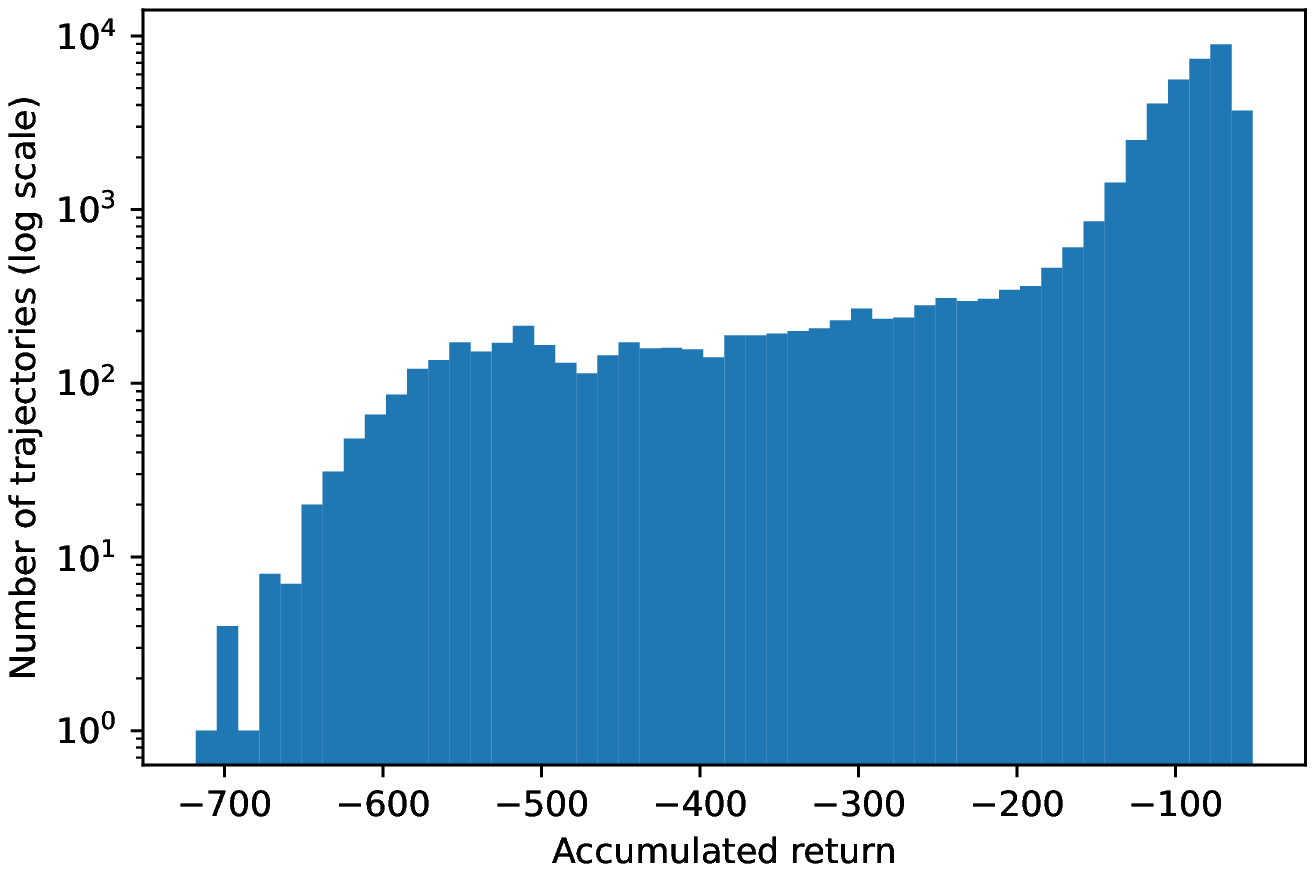}
\caption{CheetahVel}
\end{subfigure}
\caption{The histogram shows the distribution of trajectory returns within the offline data set. x axis shows the range of return and y axis shows the trajectory number counts. Note that in (a),(c), y axis are plotted with log scale.}
\label{fig: return}
\end{figure}

This work involves sampling 30 tasks as meta-train tasks and 5 tasks as meta-test tasks for each environment. In order to showcase the distribution of tasks and trajectory data, Figure~\ref{fig: goal} and~\ref{fig: return} illustrate the meta-train/test tasks and offline data qualities for three selected tasks. During the offline data collection stage, a single task learning agent is trained for each training task, and the replay buffer is saved as an offline dataset. As a result, the dataset constitutes trajectories with varying qualities measured by their accumulated returns.

\subsection*{Implementation Details}

As outlined in section~\ref{sec: framework}, the task representation learning stage involves jointly learning three components: a transition encoder that maps each one-step transition to a hidden embedding vector, an aggregator that combines all transition information and predicts the embedding for the entire trajectory context, and a projection module that projects the embeddings to a lower-dimensional space. Contrastive losses are then calculated based on the distances in the projected space. It should be noted that the projection head is discarded after the task representation learning stage.

Moving onto the policy learning stage, we modify the state space in the environments by concatenating the environmental states with predicted task embeddings, formalizing the policy learning problem as a POMDP. We then utilize IQL \citep{kostrikov2021offline} to learn policies from offline data. The key hyperparameters used in our experiments are summarized in table~\ref{tab: param1}. For further details regarding our experimental setup, please refer to our code repository at \url{https://github.com/ZJLAB-AMMI/HS-OMRL}.
\begin{table}[h]
\centering
\begin{tabular}{c c}
\hline
Parameter & Value \\
\hline
Transition encoder network & MLP$(64,64)$\\
Transition embedding dimension & $20$ \\
Aggregator network & Attention$(64,64)$\\
Aggregator output dimension & $64$\\
Projection head network & MLP$(64)$\\
Projection space dimension & $5$ \\
Contrastive batch size & $256$ \\
InfoNCE temperature $\beta$ & $0.1$\\
CL learning rate & $3e-4$\\
\hline
Actor \& Critic networks & MLP$(64,64)$ \\
RL batch size & 256\\
IQL percentile $\tau_{\text{IQL}}$& 0.8\\
IQL temperature $\beta_{\text{IQL}}$& 3.0\\
RL learning rates & $3e-4$ \\
Discount factor & $0.99$ \\
\hline
\end{tabular}
\caption{Configurations and hyperparameters used in the experiments.}
\label{tab: param1}
\end{table}
\newpage
\subsection*{Experimental results on cases without distribution shifts}
As presented in the table below, our proposed method denoted by $\mathcal{L}^{\text{HP+HG}}$, performs competitively when distribution shifts are not present. In fact, it yields the best testing performance in two experiments, namely \textit{PointRobotGoal} and \textit{WalkerParams}. However, it produces suboptimal testing performance in other experiments.
\begin{table*}[h]
    \centering
    \begin{tabular}{c c c c c c}
    \hline
        Environment & \textit{PointRobotGoal} & \textit{AntDir} & \textit{CheetahVel} & \textit{WalkerParams} & \textit{HopperParams} \\
        \hline
        FOCAL           & $-42.9\pm 8.5$        & $309.2\pm 33.6$           & $-35.6\pm 5.3$            & $225.4\pm56.4$            & $195.6\pm62.3$ \\
        FOCAL++         & $-43.3\pm8.7$         & $\mathbf{413.7 \pm 25.0}$ &  $-32.9\pm 7.4$           & $273.2\pm49.3$            & $210.6\pm37.9$ \\
        CORRO           & \emph{-23.8$\pm$4.4 }        & $354.7\pm30.7$            &  $-32.7\pm 4.6$           & \emph{301.5$\pm$37.9}            &  $247.6\pm25.6$ \\
        Offline PEARL   & $-103.7\pm11.5$       & $249.5\pm57.4$            & $\mathbf{-30.2\pm 3.3}$   & $259.1\pm48.2$            & $\mathbf{264.0\pm18.5}$\\
        $\mathcal{L}^{\text{HP+HG}}$ &
                        $\mathbf{-20.6\pm3.3}$  & \emph{409.2$\pm$27.9}            & \emph{-31.8$\pm$3.8}             & $\mathbf{312.8\pm29.7}$   & \emph{250.6$\pm$43.5}\\
        \hline
        $\mathcal{L}^{\text{SCL}}$ &
                        $-24.9\pm5.1$           & $387.2\pm31.7$            & $-33.2\pm5.1$             & $283.7\pm71.2$            & $223.7\pm31.6$\\
        $\mathcal{L}^{\text{HG}}$ &
                        $-22.8\pm5.7$           & $386.1\pm37.1$            & $-34.5\pm3.0$             & $292.4\pm43.8$            & $257.0\pm26.8$\\
        $\mathcal{L}^{\text{HP}}$ &
                        $-25.3\pm6.8$           & $389.2\pm37.2$            & $-32.0\pm4.2$             & $285.9\pm52.8$            & $229.9\pm41.2$ \\
        \hline

    \end{tabular}
    \caption{Performance comparison in terms of accumulated returns, for cases without distribution shifts. The best and suboptimal results are marked in \textbf{bold} and \emph{italics}, respectively. All results are averaged over 5 random seeds.  }
    \label{tab: high}
\end{table*} 
\end{document}